\documentclass[sigconf]{aamas}

\usepackage{balance} 
\usepackage{amsmath}
\usepackage{caption}
\usepackage{fix-cm}
\usepackage{microtype}
\usepackage{textcomp}
\usepackage{graphicx}
\usepackage{xcolor}
\makeatletter
\renewcommand\footnotetextcopyrightpermission[1]{}
\makeatother

\makeatletter
\newcommand\blfootnote[1]{%
  \begingroup
  \renewcommand\thefootnote{}%
  \def\Hy@raisedlink##1{}%
  \footnotetext{#1}%
  \endgroup
}
\makeatother

\setcopyright{none}
\copyrightyear{}
\acmYear{}
\acmDOI{}
\acmISBN{}

\title[AAMAS-2027 Formatting Instructions]{Collaborative Memory for Multi-Agent VLM Systems}

\author{Huixin Zhang\texorpdfstring{\textsuperscript{*}}{}}
\affiliation{
  \institution{Texas A\&M University}
  \city{College Station}
  \country{United States}}
\email{zhanghui21@tamu.edu}

\author{Shao-Jun Xia%
\texorpdfstring{\textsuperscript{*}\textsuperscript{\textdagger}}{}}
\affiliation{
  \institution{Duke University}
  \city{Durham}
  \country{United States}
}
\email{shaojun.xia@duke.edu}

\author{Di Wang%
\texorpdfstring{\textsuperscript{\textdagger}}{}}
\affiliation{
  \institution{Foxconn}
  \city{Milwaukee}
  \country{United States}
}
\email{di.wang@foxconn.com}

\author{Liangxi Liu%
\texorpdfstring{\textsuperscript{\textdagger}}{}}
\affiliation{
  \institution{Northeastern University}
  \city{Boston}
  \country{United States}
}
\email{liu.liangx@northeastern.edu}

\author{Hainan Xiong}
\affiliation{
  \institution{Harvard University}
  \city{Cambridge}
  \country{United States}}
\email{hainanxiong@alumni.harvard.edu}

\author{Zihao Wang}
\affiliation{
  \institution{Meta}
  \city{Menlo Park}
  \country{United States}}
\email{wangzihao1@g.ucla.edu}

\begin{abstract}
Vision-language model (VLM) agents combine specialized perception, tools, and reasoning to address complex visual tasks. In multi-agent settings, different agents inspect different image regions, video frames, or visual representations, so collaboration extends beyond distributed reasoning to distributed perception. This makes shared visual context a central problem in VLM agent collaboration. In this paper, we frame memory hierarchy, cross-agent sharing, and consistency mechanisms around the need to reconcile interpretations and update dependent reasoning. Effective collaboration requires agents to build on contributions from other agents, recover missing visual context, and reconcile differing interpretations as new evidence emerges. Shared visual memory preserves not only images or textual summaries but also the dependencies among observations, agent interpretations, and subsequent reasoning. Together, these design considerations shape how information flows and evolves across VLM agents. The proposed framework provides a foundation for building reliable and resource-efficient agent teams.
\end{abstract}


\newcommand{\BibTeX}{\rm B\kern-.05em{\sc i\kern-.025em b}\kern-.08em\TeX}

\begin{document}

\fancypagestyle{standardpagestyle}{%
  \fancyhf{}%
  \fancyfoot[C]{\thepage}%
  \renewcommand{\headrulewidth}{0pt}%
  \renewcommand{\footrulewidth}{0pt}%
}

\fancypagestyle{firstpagestyle}{%
  \fancyhf{}%
  \fancyfoot[C]{\thepage}%
  \renewcommand{\headrulewidth}{0pt}%
  \renewcommand{\footrulewidth}{0pt}%
}

\pagestyle{standardpagestyle}

\makeatletter
\ifdefined\lx@clear@frontmatter
  \lx@clear@frontmatter{ltx:date}[role=copyright]
\fi
\makeatother
\maketitle

\blfootnote{%
  \begin{tabular}{@{}l@{~}l}
    \textsuperscript{*} & Equal contribution.\\
    \textsuperscript{\textdagger} & Corresponding authors.
  \end{tabular}%
}

\section{Introduction}
\label{sec:intro}

A vision-language model (VLM) integrates visual observations with textual instructions~\cite{liu2023visual,li2023blip}. A VLM-based agent extends this multimodal capability with task context, external tools, and auxiliary models, enabling goal-directed perception and reasoning~\cite{shen2023hugginggpt}. We argue that collaboration among such agents introduces a distinct systems challenge: intermediate visual observations and interpretations must remain accessible, verifiable, and revisable across agent boundaries. This becomes particularly important when agents use different task contexts and tools to divide perceptual and reasoning responsibilities.
Such collaboration requires more than exchanging answers because agents need to discover what others have observed, obtain missing evidence, and revisit conclusions when earlier interpretations change.
\emph{How should multiple VLM agents divide visual work and manage shared memory?} The goal is to preserve intermediate observations and interpretations together with the context needed for other agents to reuse, verify, and revise them.

Work on multi-agent systems reports gains in tasks requiring complementary visual strengths, such as mobile interface operation and long-video reasoning~\cite{wang2024mobile,chen2026videochat,wang2026think}, and coordination becomes particularly demanding with continuous input or high-resolution sources.
Streaming video adds a need for timely responses alongside ongoing observation and access to earlier context~\cite{chen2024videollm,xu2026streamingvlm,qian2025dispider,tang2026asynchronous}. Pathology review can combine local detail from gigapixel slides with wider tissue context and associated reports~\cite{xu2024whole,ding2025multimodal}.
Such settings motivate teams that inspect different regions, intervals, or subtasks while sharing the evidence needed to integrate and check their findings.

Shared memory provides a substrate for connecting observations, interpretations, and reasoning across agents. Prior work on memory for large language model (LLM) agents has examined how information is retained, retrieved, organized, and revised~\cite{zhang2025survey,hu2025memory}. At the single-agent level, existing systems focus on context management and structured memory records~\cite{packer2023memgpt,xu2026mem}, while recent work has begun to study selective memory sharing and reuse across agents and models~\cite{rezazadeh2025collaborative,chang2026memcollab}. For VLM teams, however, collaboration introduces an additional requirement. Observations and interpretations must remain linked to the visual context that enables other agents to reuse, verify, and, when necessary, revise them.

\begin{figure}[t]
\centering
\resizebox{0.958\linewidth}{!}{%
\includegraphics[trim=1pt 1pt 1pt 1pt,clip]{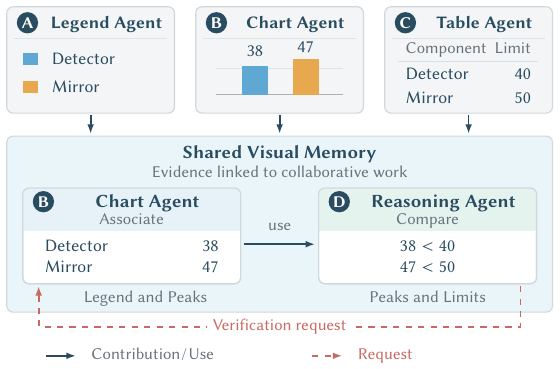}
}
\caption{\small \textbf{Collaboration through Shared Visual Memory.} \mdseries
The chart agent combines extracted values with labels provided by the legend agent, while the reasoning agent compares the labeled values with limits from the table agent. When verification is requested, the chart agent reexamines the retained bars and legend and revises the association if necessary.}
\Description{The legend, chart, and table agents contribute complementary observations. The chart agent uses the legend to associate temperatures with components. The reasoning agent compares the associated peaks with limits from the table agent and can request verification. The chart agent then reexamines the retained bars and legend before revising the association.}
\label{fig:motivation}
\vspace{-12pt}
\end{figure}

Visual tasks create a tension between specialization and shared understanding. Chart and infographic tasks require interpreting graphical elements together with labels and layout~\cite{masry2022chartqa,mathew2022infographicvqa}. An agent can work efficiently on a small region, but that region can omit a legend needed by a collaborator. Revising a bar-to-label association can invalidate a comparison by another agent without changing pixels.
Information sufficient for local work can therefore be insufficient for collaborators to use or check a contribution.

We therefore propose organizing shared visual memory around the connections among observations, interpretations, and subsequent reasoning across agents. A memory hierarchy determines what context is retained, cross-agent sharing governs access to usable representations, and consistency mechanisms make revisions visible to agents using earlier interpretations. These design decisions are interdependent because later verification can require additional context that memory must retain and make accessible.

We develop this design through a spacecraft thermal-testing example and three questions. How should memory link local observations to the context collaborators need? How should access adapt to the task and model interface of a receiving agent? How should a revision prompt collaborators to review dependent conclusions?

\section{A Collaborative View of Shared Visual Memory}
\label{sec:motivation}

This section uses a spacecraft thermal-vacuum test scenario, in which hardware is assessed under controlled pressure and temperature, to expose the core requirements of shared visual memory. A reported test processed terabytes of raw data~\cite{kimble2016cryo}, and its temperature histories, instrument diagrams, and requirement tables can exceed the context window of a single agent. When anomaly investigations must finish by a fixed deadline while new measurements keep arriving, agents can inspect different components and intervals in parallel, and completed analyses are then passed to a reasoning agent. Separate working contexts ensure each agent focuses on the domain knowledge its specialized tools need.

\vskip 0.5em
\noindent\textbf{\textit{Complementary observations.}}
A shared task can require agents to combine observations that are insufficient in isolation.
Legend agent A combines a model for engineering drawing understanding with sensor location records to associate plotted colors with components. Chart agent B uses calibration records, time-series analysis tools, and a thermal simulation model to interpret temperature peaks and gradients. Table agent C uses test procedures and table extraction tools to select limits for the current test phase and hardware configuration. Reasoning agent D combines these contributions and requests further inspection.
In the example, B needs the legend from A; B and C can initially work in parallel, but the comparison by D requires information to pass between the tasks. These intermediate results form a dependency structure in which later reasoning builds on earlier observations and interpretations.

Figure~\ref{fig:motivation} abstracts this workflow to a single interval using a light-sensitive detector and an optical mirror, with peak temperatures of 38 and 47 kelvin against respective limits of 40 and 50 kelvin.

\vskip 0.5em
\noindent\textbf{\textit{Iterative collaboration.}}
Shared memory allows the analysis to develop through repeated interactions.
B can publish values with unresolved component labels, request the legend, and inspect the visual evidence saved by A before updating the association.
D can compare the labeled values with the limits from C and issue a verification request if the interpretation remains uncertain. B can revise the association after the check, prompting D to reconsider the comparison.
Agents therefore alternate between contributing, reading, requesting, and reviewing.
This division of work can follow source access, tool capabilities, or expertise, as in configurable language-agent teams~\cite{wu2024autogen}. Roles can change during the task, and agents can use the same or different VLMs. A coordinator, or the agents themselves, can consult shared records of completed work and outstanding requests to decide what to do next.

\vskip 0.5em
\noindent\textbf{\textit{Shared records of evidence and work.}}
We view shared memory as a structured coordination substrate through which VLM agents can selectively access, update, and retain task-relevant evidence, interpretations, and state. Within a task, it supports a shared working memory for ongoing collaboration; across tasks, selected records persist for later reuse,
with the links needed to recover context and revise prior interpretations.

\vskip 0.5em
\noindent\textbf{\textit{Task-dependent visual context.}}
An observation retained in isolation can lose the context needed for interpretation.
For example, B needs the link between a bar and a legend entry, just as table values need cells and events need surrounding frame sequences. A memory record can therefore span several related regions or observations.
Yet providing every collaborator with every full-resolution source increases communication and processing costs, a concern also motivating selective visual retrieval~\cite{wu2025visual}. A textual interpretation supports routine reasoning, while linked visual context enables a collaborator to extend or check the interpretation. Contextual dependencies also arise across pages, images, and video segments~\cite{dong2026benchmarking,wang2025muirbench,yeo2026worldmm}, even when each local observation is understood correctly.

\section{Memory Organization and Sharing}
\label{sec:framework}

\begin{figure*}[t]
\centering
\resizebox{0.985\linewidth}{!}{%
\includegraphics[trim=1pt 1pt 1pt 1pt,clip]{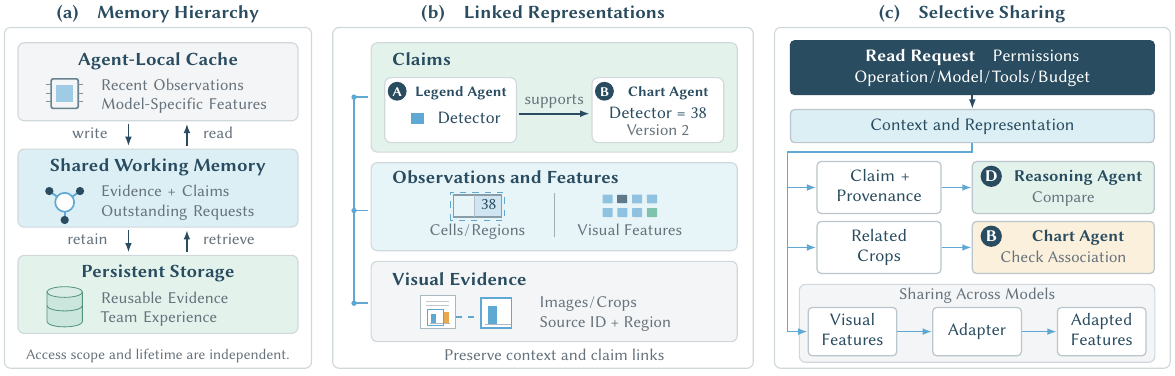}
}
\caption{
\small
\textbf{Memory Organization, Representation, and Access.} \mdseries
\textbf{(a)}~Local caches, shared working memory, and persistent storage separate access scope from retention lifetime.
\textbf{(b)}~The legend claim supports the revised chart record in Version 2, linked to observations and visual evidence.
\textbf{(c)}~Selective sharing adapts context and representation to the receiving agent.
Adapters support feature reuse across models.}
\Description{
(a)~The proposed hierarchy comprises local caches, shared working memory, and persistent records.
(b)~The legend agent associates blue with Detector. This interpretation supports the revised chart interpretation, Detector equals 38, labeled Version 2. Interpretations remain linked to observations, features, and visual sources.
(c)~A reasoning agent receives a claim with provenance for comparison, while a chart agent receives related crops to check the association. A separate model-adaptation path maps visual features into adapted features for the receiving model.
}
\label{fig:framework}
\vspace{-7.95pt}
\end{figure*}

\noindent\textbf{\textit{Memory hierarchy.}}
A VLM team needs fast access to current observations as well as selective reuse of earlier work.
Figure~\ref{fig:framework} outlines a memory hierarchy with three levels: agent-local caches for recent observations and model-specific features; shared working memory for evidence, claims, and outstanding requests; and persistent storage for subsequent tasks. Access scope and retention lifetime are separate choices. Persistent records can also be private, and the illustrated stores need not be physically separate.

In streaming video~\cite{qian2025dispider,tang2026asynchronous}, one agent could cache recent frames and publish event descriptions linked to retrievable, timestamped clips, while other agents inspect earlier clips or integrate the shared findings. Coordination also requires shared working memory to distinguish inspected intervals from those awaiting inspection, so that delayed processing is not mistaken for an absence of events.

\vskip 0.5em
\noindent\textbf{\textit{Linking representations.}}
Within the hierarchy, source images and crops retain visible evidence, so that the team need not rely entirely on textual descriptions~\cite{nguyen2026personal,su2026history,xia2026memory}.
Visual features and structured observations provide processed representations, while claims summarize interpretations made by agents. Visual features are vectors produced by a visual encoder. Structured observations include recognized text with bounding boxes, table cells, and associations between regions.
These representations preserve different information rather than forming a universal ordering of fidelity. A crop can omit a distant legend, while a concise structured record can completely describe a particular comparison.

Linking the representations allows an agent to move from a compact claim to the context needed for verification. Provenance records the source region, contributing agent, and prior observations cited by an interpretation. For example, the claim that the detector peaked at 38 kelvin needs links to both the bar and the legend association. Stable source identifiers and coordinates support retrieval, while explicit links between interpretations support subsequent revision.

\vskip 0.5em
\noindent\textbf{\textit{Selective sharing.}}
The amount and form of context to share depend on the work assigned to the receiving agent. For example, given the upper limit from C, D can check a temperature using the component, test phase, interval, peak value, unit, and source reference from B. Checking the association instead requires B to inspect the bars and legend together. The sharing policy therefore selects related evidence and a representation suited to the current operation, model interface, available tools, and budget of the receiving agent.
The access interface also identifies the representation, the source, and the interpretation version being read, together with the permissions required to retrieve the evidence.

\vskip 0.5em
\noindent\textbf{\textit{Sharing across models.}}
Visual exchange and adapters support communication beyond natural language~\cite{yu2026visual,chen2026scaling}. Vision Wormhole~\cite{liu2026vision} conveys latent reasoning through visual input interfaces using encoders and decoders for different VLM families. Reusing encoded visual features can reduce repeated visual processing, but distinct VLMs differ in feature dimensions, positional encodings, and input interfaces.
For feature reuse, one possible approach is an adapter that maps stored features into the input space of the receiving model while leaving the shared representation fixed. If an interface does not expose feature inputs, the receiving agent must use an image or another representation.
Such cross-model feature reuse must preserve links to the visual sources beyond matching tensor dimensions. The value of reuse depends on collaborative reasoning quality and total costs of adaptation, encoding, retrieval, and repeated inspection, not on compatibility or compression alone.

\begin{figure*}[t]
\centering
\resizebox{\linewidth}{!}{%
\includegraphics[trim=1pt 1pt 1pt 1pt,clip]{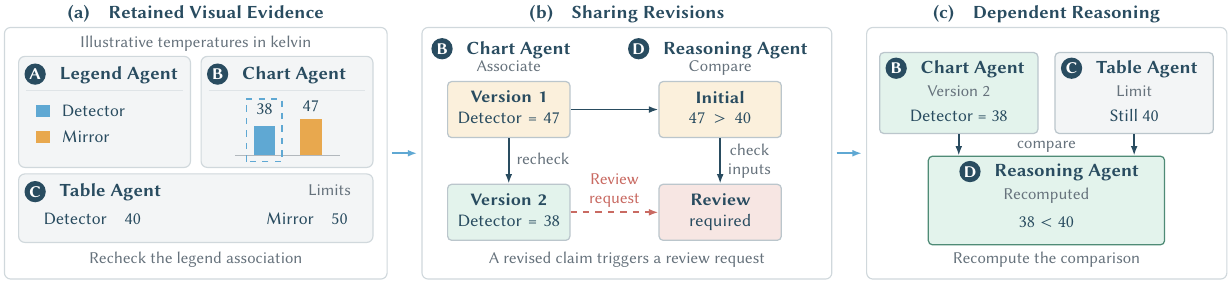}
}
\caption{\small
\textbf{Claim Revision and Dependent Reasoning.}
\mdseries
\textbf{(a)}~The chart agent rechecks bars against the legend.
\textbf{(b)}~A revised claim prompts a review request. The reasoning agent rechecks the earlier input.
\textbf{(c)}~The reasoning agent recomputes the flag from the revised value and unchanged limit.
Solid arrows denote use or processing; the dashed arrow denotes the review request.
}
\Description{
(a)~The chart agent rechecks the bars against the retained legend.
(b)~A revised association prompts a review request. The reasoning agent checks whether the input to the earlier comparison has changed.
(c)~The reasoning agent rechecks the temperature flag using the revised value and unchanged limit. Solid arrows denote use or processing, while the dashed arrow denotes the review request.
}
\label{fig:consistency}
\vspace{-11.5pt}
\end{figure*}

\section{Consistency and Reconciliation}
\label{sec:consistency}

\noindent\textbf{\textit{Reconciling interpretations.}}
Consistency in collaborative VLM systems depends on whether agent interpretations are grounded in compatible evidence, source states, and contextual assumptions. Interpretation mismatches can arise when agents inspect different image regions, video frames, or source versions, so reconciliation must establish whether conflicting claims are actually conditioned on the same evidence and context. Agreement alone does not constitute independent confirmation, since multiple agents can inherit the same erroneous interpretation~\cite{wang2026seeing,lin2026beyond}. Shared memory should therefore preserve the evidence examined by each agent and the interpretations derived from it, enabling conflicts to be resolved through evidence-level reconciliation and targeted reinspection.

\vskip 0.5em
\noindent\textbf{\textit{Revising dependent reasoning.}}
In the example, B initially assigns the mirror peak of 47 kelvin to the detector, leading D to flag a violation of the 40 kelvin limit. Rechecking the legend from A against the saved bars associates the detector with 38 kelvin (see Figure~\ref{fig:consistency}).
If B revises the association, D needs to recompute the comparison using the unchanged limit from C. Without recomputation, shared memory can contain a revised association alongside a conclusion based on the earlier version.

Truth maintenance has long used recorded reasons to support belief revision~\cite{doyle1979glimpse}. Handling outdated information is also an established concern in conversational memory~\cite{chhikara2025mem0,wu2025longmemeval}. Keeping earlier and revised records distinguishable and passing corrections between agents provide an existing basis for addressing revisions~\cite{feng2026diachronic}. For visual tasks, revision must also track associations that can change while the source image and extracted numbers remain unchanged. We therefore propose linking derived claims to supporting interpretations so that contributing agents can request a review of dependent conclusions. Such a request identifies work requiring review but does not establish the replacement answer.

\vskip 0.5em
\noindent\textbf{\textit{Sharing revisions.}}
Revision also concerns when collaborators learn of a change. For example, if D is already reasoning from an earlier interpretation by B, updating the shared record does not update the local context of D. Before publishing a result, D can check whether the stored claims still match those used in the comparison. Concurrent, conflicting claims should remain distinguishable until examined, rather than silently overwriting one another.

\vskip 0.5em
\noindent\textbf{\textit{Reliability and synchronization.}}
Waiting for every observation to be verified can stall collaboration, while proceeding with unverified claims can spread errors. A practical policy must balance delay against error propagation while preserving alternative interpretations and making revisions visible. Recorded dependencies support selective review, but missing links can leave affected conclusions unexamined, so incomplete dependencies call for broader reinspection. Recording the inputs delivered to an agent also does not establish which inputs the model actually used. A consistent set of records can still be wrong, and verification itself can introduce mistakes. The objective is reliable joint reasoning without repeating every observation across agents.

\section{Related Work}
\label{sec:related}

\noindent\textbf{\textit{Multi-agent memory and coordination.}}
Blackboard systems coordinate specialized problem solvers through shared intermediate results~\cite{nii1986part}. INMS~\cite{gao2024inms} studies language-agent memory sharing, G-Memory~\cite{zhang2026g} organizes team experience hierarchically, and DecentMem~\cite{hao2026self} supports collaboration while keeping experience memories private to each agent. Yu et al.~\cite{yu2026multi} analyze memory hierarchy, cross-agent cache sharing, access protocols, and consistency from a computer architecture perspective. MAGE~\cite{feng2026diachronic} extends memory curation to multimodal records, orchestration, provenance, and revision, retaining links to original sources alongside textual descriptions. Surveys also identify role-aware and learned shared-memory management as open directions~\cite{hu2025memory}.
We build on these foundations to further address how visual evidence, interpretations, and dependent reasoning remain linked across collaborating agents.

\vskip 0.5em
\noindent\textbf{\textit{Collaboration among visual agents.}}
ViF~\cite{yu2026visual}, dual latent memory~\cite{yu2026dual}, and compact latent collaboration~\cite{chen2026scaling} explore visual or latent exchange between agents, including shared perception and heterogeneous interfaces. AgentDet~\cite{li2026agentdet} provides shared and persistent visual knowledge for specialized roles, while MATA~\cite{cai2026mata} learns agent transitions from shared execution memory. ORCA~\cite{lassoued2026orca} coordinates document specialists, and MARDoc~\cite{chen2026mardoc} links evidence to reasoning nodes in memory refined by an iterative document-analysis team. EAGLE~\cite{wang2026seeing} and CAMA~\cite{lin2026beyond} address visual consensus and memory arbitration, respectively. These systems demonstrate the value of visual exchange, shared execution state, and evidence-aware coordination, and they motivate us to examine how such information should be organized into a persistent and revisable memory structure across agents.

\vskip 0.5em
\noindent\textbf{\textit{Visual retention and reuse.}}
GraphMemix~\cite{li2026graphmemix} organizes multimodal evidence for retrieval, AdaMM~\cite{tian2026beyond} supports analysis over retained multimodal content, and SlideBank~\cite{zhao2026slidebank} studies persistent hierarchical visual evidence. Benchmarks also study memory over long multimodal histories and evidence spread across sources~\cite{ren2026memlens,chai2026smmbench}.
For collaboration, an agent needs the context to build on an observation, a usable representation, and a way to identify reasoning affected by a revised association. These requirements connect visual retention to team organization and reliability.

\section{Future Directions}
\label{sec:agenda}

A longer-term direction is to build VLM teams that can resume unfinished analyses and reuse experience across tasks. A newly joined agent needs access to retained observations, current interpretations, and pending requests, along with their provenance and relevant conditions.

\vskip 0.5em
\noindent\textbf{\textit{Memory and task allocation.}}
Changes in evidence needs or available tools create a need to reconsider who inspects which evidence and what context collaborators receive. A central question is how to jointly learn task allocation and the selection of observations, interpretations, and unresolved requests for shared memory. Adapting task allocation and memory selection together should be compared against fixed roles that request additional context. Feedback should combine correctness checked against task reference answers with communication and processing costs.

\vskip 0.5em
\noindent\textbf{\textit{Learning from repeated collaboration.}}
Experience-based learning in language agents~\cite{shinn2023reflexion,zhao2024expel} suggests distilling reusable procedures from successful and failed attempts. Retained experience should link each procedure to its supporting evidence, independently checked outcomes, and applicable conditions. Later corrections should prompt review of the experience derived from those outcomes. Evaluation on new tasks then tests whether a procedure transfers beyond the cases that formed it.
The target is correct revisions with less repeated processing or delay, accounting for maintenance costs. Short tasks, rare revisions, and changes affecting most conclusions test when maintenance outweighs the gains.

\section{Conclusion}

VLM agent teams distribute perception as well as reasoning, making shared visual context essential to collaboration. In this paper, we jointly design the memory hierarchy, cross-agent sharing, and consistency mechanisms to maintain dependencies across observations, interpretations, and downstream reasoning. The framework establishes a foundation for reliable and resource-efficient VLM agent teams, enabling effective knowledge retention and coherent reasoning.

\clearpage
\bibliographystyle{unsrtnat}
\bibliography{ref}

\end{document}